Research

# Echo-DND: a dual noise diffusion model for robust and precise left ventricle segmentation in echocardiography

**Abdur Rahman[1] · Keerthiveena Balraj[2] · Manojkumar Ramteke[1,2] · Anurag Singh Rathore[1,2]**

**Abstract**
Recent advancements in diffusion probabilistic models (DPMs) have revolutionized image processing, demonstrating significant potential in medical applications. Accurate segmentation of the left ventricle (LV) in echocardiograms is crucial for diagnostic procedures and necessary treatments. However, ultrasound images are notoriously noisy with low contrast and ambiguous LV boundaries, thereby complicating the segmentation process. To address these challenges, this paper introduces Echo-DND, a novel dual-noise diffusion model specifically designed for this task. Echo-DND leverages a unique combination of Gaussian and Bernoulli noises. It also incorporates a multi-scale fusion conditioning module to improve segmentation precision. Furthermore, it utilizes spatial coherence calibration to maintain spatial integrity in segmentation masks. The model's performance was rigorously validated on the CAMUS and EchoNet-Dynamic datasets. Extensive evaluations demonstrate that the proposed framework outperforms existing state-of-the-art models. It achieves high Dice scores of 0.962 and 0.939 on these datasets, respectively. The proposed Echo-DND model establishes a new standard in echocardiogram segmentation, and its architecture holds promise for broader applicability in other medical imaging tasks, potentially improving diagnostic accuracy across various medical domains.

**Article highlights**
The key contributions are as follows:

- We introduce Echo-DND, a dual noise diffusion model, which seamlessly integrates both Gaussian and Bernoulli noises in a singular diffusion pipeline.
- We propose a multi-scale fusion-based efficient conditioning technique that uses multi-resolution feature extraction and cross-resolution fusion for enhanced boundary delineation.
- Our training method incorporates a pixel-wise spatial coherence calibration technique that complements the diffusion-based segmentation process, ensuring superior segmentation outcomes.

✉ Anurag Singh Rathore, asrathore@biotechcmz.com | [1]Department of Chemical Engineering, Indian Institute of Technology Delhi, Hauz Khas, New Delhi 110016, India. [2]Yardi School of Artificial Intelligence, Indian Institute of Technology Delhi, Hauz Khas, New Delhi 110016, India.

## 1 Introduction

Echocardiography, as the most widely used imaging modality for examining heart diseases, plays a pivotal role in clinical cardiology [1, 2]. By providing real-time images of cardiac structures, echocardiograms enable clinicians to locate abnormalities, assess cardiac function, and evaluate critical metrics such as ejection fraction. The latter, calculated by measuring the change in left ventricular volume between end-diastole and end-systole, serves as a quantitative indicator in diagnosing cardiac dysfunction. It is typically done manually by experts. However, the manual tracing of endocardial borders is error-prone, time-consuming, and subject to inter-observer variability and biases. The advent of deep learning architectures, particularly Convolutional Neural Networks (CNNs) and transformer-based models, has revolutionized medical image analysis [3]. These models have achieved benchmark results across various tasks, including left ventricular (LV) segmentation in echocardiograms. Despite these advancements, echocardiograms present unique challenges. Their low-contrast images, coupled with the complex geometry of ventricles and atria, introduce boundary ambiguity. Consequently, existing models still fall short of achieving optimal performance in this domain.

Recently, diffusion probabilistic models (DPMs) have emerged as powerful generative models capable of producing high-quality images with remarkable diversity [4, 5]. Traditionally applied to generate new images, DPMs have recently found utility in image segmentation tasks [6, 7]. Their potential lies in their ability to yield more precise segmentation masks by leveraging the underlying noise distribution during the diffusion process and, hence, effectively modeling the inherent uncertainty associated with it. Usually, for diffusion models, Gaussian noise is the preferred choice for its mathematical tractability, well-defined statistical properties, and accurate representation of physical phenomena such as sensor noise and thermal noise [8]. However, we posit that incorporating alternative noise types can enhance the adaptability and performance of these models. For binary segmentation tasks like left ventricular segmentation in echocardiograms, a noise distribution that closely aligns with the underlying binary nature of the task can better capture inherent uncertainties. Thus, we propose a dual noise strategy in our Echo-Dual Noise Diffusion (DND) model, integrating both Gaussian and Bernoulli noise. This approach allows Gaussian noise to model continuous variations and sensor noise, while Bernoulli noise captures the binary nature of segmentation masks. By combining these noise types, Echo-DND achieves more accurate and robust segmentation results, leveraging the complementary strengths of each noise distribution.

Addressing the limitations of existing diffusion-based segmentation techniques, our approach incorporates a multi-scale fusion-based conditioning mechanism. Traditional CNN models [9–12] used for feature extraction often rely on either high or low resolutions, which can miss crucial details or fail to capture the broader context. By simultaneously upsampling and downsampling features across multiple scales and fusing them, our method achieves a comprehensive understanding of the image, accurately capturing intricate boundary details. This approach has been successfully used in [13] to enhance recognition accuracy. Additionally, integrating a calibration technique to refine the outputs of the model can further enhance performance. While diffusion models excel at capturing fine-grained details via pixel-wise uncertainty modeling, they sometimes struggle with maintaining pixel-wise spatial relationships due to their iterative nature. Conversely, traditional CNN-based feedforward styles employ cross-entropy loss functions [14], enabling them to effectively capture pixel-wise spatial relationships but potentially lacking detail. By incorporating a CNN-based feedforward style cross-entropy loss into the diffusion process, a synergy is achieved. The CNN complements the fine-grained details captured by diffusion, resulting in improved segmentation quality. The proposed work purposefully addresses these limitations and is engineered specifically for LV segmentation in echocardiograms.

Our approach significantly surpasses existing methods, establishing a new benchmark in LV segmentation tasks. The innovative design of the Echo-DND model paves the way for more accurate and reliable cardiac diagnostics, offering substantial improvements in clinical practice. The remainder of this paper is organized as follows: Section 2 reviews existing methodologies in echocardiographic segmentation, including traditional, CNN-based, transformer-based, and recent diffusion-based methods. Section 3 provides a foundational overview of standard diffusion models and the broader context of image segmentation. Section 4 details the technical aspects of the Echo-DND model. Section 5 presents the experimental setup, datasets used, and a comparative analysis of the results. Finally, Section 6 summarizes the contributions and impact of the Echo-DND model on echocardiographic segmentation and clinical practice.

## 2 Related work

Early efforts in left ventricular (LV) segmentation employed diverse mathematical, classical machine learning, and image processing techniques to delineate LV borders with modest accuracy [15–18]. These methods laid the foundational groundwork for automated segmentation but were often limited by their accuracy and computational efficiency. The advent of deep learning heralded a paradigm shift in LV segmentation. Convolutional Neural Networks (CNNs), particularly architectures based on U-Net [11], emerged as the cornerstone of this transformation. The U-Net model was noted for its first contracting and then expanding pathways, along with skip connections, which enabled effective multiscale feature extraction from images. This architecture was further refined in various variants like nnU-Net [19], U-Net++ [20], Res-UNet [21], MADR-Net [22], CR-Unet [23] and ResDUnet [24], which enhanced segmentation performance by introducing improvements such as dense connections and residual learning. DeepLabV3 [55] introduced atrous convolutions within a pyramid structure, effectively addressing object scale variations in echocardiograms. Beyond U-Net variants, other successful deep learning approaches include LU-Net [25], a two-stage network inspired by Mask R-CNN [26]. LU-Net first predicts a region of interest (ROI) around the heart, followed by high-accuracy segmentation within the ROI. ResDUnet [24] improved upon U-Net by integrating cascaded dilated convolutions for multi-scale feature extraction, whereas MFP-Unet [27] incorporated FPNs to enhance information flow within the U-Net architecture. Further combining effective components, Azizi et al. [51] developed a hybrid model utilizing the U-Net encoder enhanced with Atrous Spatial Pyramid Pooling and the efficient DeeplabV3+ decoder, achieving strong results on echocardiography segmentation.

Beyond architectural improvements, recent advancements have explored novel avenues to overcome challenges in medical image analysis. The scarcity of annotated data remains a significant bottleneck, prompting the emergence of contrastive pre-training methods that leverage self-supervised learning to achieve competitive segmentation with limited labeled data [27, 28]. Similarly, tackling sparse annotations specifically in video data, Maani et al. [52] proposed SimLVSeg, which employs self-supervised temporal masking for pre-training followed by weakly-supervised fine-tuning of video segmentation networks, effectively utilizing unlabeled frames. Additionally, addressing temporal consistency across cardiac cycles has become paramount. Innovations like interpretable autoencoders [29] and multi-view temporal models such as MV-RAN [30] have demonstrated promise in enhancing LV segmentation accuracy by leveraging temporal and multi-view information.

Vision Transformers (ViTs) [31], leveraging attention mechanisms [32] for image recognition, have recently shown potential in medical image segmentation. Methods like TransBridge [33], TransUNet [34], SegFormer [35], and Swin Transformer [12], BAT-Former [36], HST-MRF [37] and SSCFormer [38] have achieved state-of-the-art results by adapting transformer architectures to segmentation tasks. Meanwhile, Generative Adversarial Networks (GANs) [39], renowned for image generation, have also been applied to medical image segmentation, including echocardiograms [40–42]. However, recent advancements suggest that diffusion models offer superior fidelity in image synthesis [4]. Diffusion models operate by introducing noise into an image in a controlled manner and progressively denoising it to recover the original image, a process that can be reversed to generate realistic new images. This makes them particularly attractive for segmentation tasks, where the goal is to iteratively refine a noisy representation of the target object (e.g., the LV) into a precise segmentation mask. Building on this promise, recent works like EnsemDiff [43], SegDiff [6], BerDiff [44] and MedSegDiff [7] have explored diffusion models for image segmentation, achieving promising results.

Building upon these advancements, particularly [7], we propose Echo-DND, a novel diffusion-based approach tailored for LV segmentation in echocardiograms. Echo-DND integrates a multi-scale conditioning system, dual noise diffusion strategy, and calibration technique to achieve superior segmentation accuracy, particularly in challenging imaging conditions.

## 3 Background

### 3.1 Gaussian diffusion

Diffusion models [5, 45] are a powerful category of generative models that excel at producing high-quality data by reversing a predefined stochastic process. These models work by gradually converting structured data into noise through a forward diffusion process and then learning to invert this process to generate new data (Fig. 1). The forward diffusion process begins with the input data and incrementally adds Gaussian noise at each timestep, resulting in a sequence of progressively noisier latent variables $x_1, x_2, \ldots, x_T$. Mathematically, this process (as presented in DDPM [5]) can be described as a Markov chain that progressively adds noise to the input data:

$$q(x_{1:T}|x_0) = \prod_{t=1}^{T} q(x_t|x_{t-1}) \tag{1}$$

where the conditional probability distribution $q(x_t|x_{t-1})$ is given by:

$$q\left(x_t|x_{t-1}\right) = \mathrm{N}\left(xt; \sqrt{1-\beta_t}x_{t-1}, \beta_t I\right). \tag{2}$$

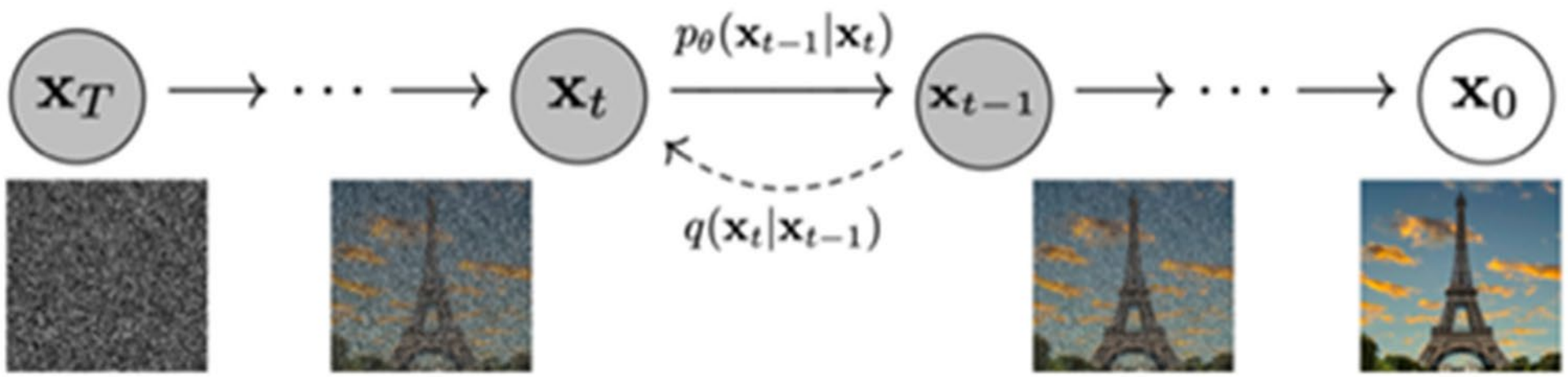

**Fig. 1** Visualization of the reverse diffusion process, transforming Gaussian noise into an image of the Eiffel Tower, illustrating the model's generative capabilities

Here, $x_0$ is the original clean segmentation mask, $x_t$ is the noised segmentation mask at timestep $t$, $\beta_t$ is a variance schedule that dictates the amount of noise added at each timestep $t$, $N(\mu, \Sigma)$ denotes a normal distribution with mean μ and covariance Σ, and I represents the identity matrix. As $t$ progresses, the data becomes increasingly corrupted until, after $T$ steps, it converges to $x_T$ which approximates a sample from a standard Gaussian distribution. To generate new data, diffusion models learn a reverse process that denoises the corrupted data. This reverse process is parameterized by a neural network $\epsilon_\theta$ and described by:

$$p_\theta\left(x_0\right) = p\left(x_T\right) \prod_{t=1}^{T} p_\theta\left(x_{t-1}|x_t\right). \tag{3}$$

This involves learning the distributions $p_\theta(x_{t-1}|x_t)$ for the reverse transitions, modeled as:

$$p_\theta\left(x_{t-1}|x_t\right) = N\left(x_{t-1}; \mu_\theta\left(x_t, t\right), \Sigma_\theta\left(x_t, t\right)\right). \tag{4}$$

In this context, $\mu_\theta\left(x_t, t\right)$ signifies the predicted mean, and $\Sigma_\theta(x_t, t)$ denotes the predicted variance for the given parameters. As outlined in the foundational DDPM work [5], the U-Net model ϵθ is used to predict the noise embedded within the input. The entire training procedure for the network aims to minimize the discrepancy between the true forward process and the learned reverse process, ensuring accurate denoising and high-quality data generation.

### 3.2 Bernoulli diffusion

Expanding on the principles of Gaussian diffusion models, the Bernoulli diffusion process, built upon the foundational diffusion principles [50], offers a specialized approach for handling discrete data, particularly in the context of binary segmentation masks, and has been applied in works like BerDiff [44]. Unlike Gaussian diffusion, which introduces continuous noise, Bernoulli diffusion applies a forward process that adds Bernoulli noise at each timestep. This results in a sequence of progressively noisier binary variables, with the forward process governing the transition from $x_{t-1}$ to $x_t$ encapsulated by $q(x_t|x_{t-1})$, given by:

$$\mathrm{q}\left(x_t|x_{t-1}\right) = Bernoulli\left(xt; \left(1-\beta_t\right)x_{t-1} + \frac{\beta_t}{2}\right).$$

(5It introduces a novel conditioning mec)

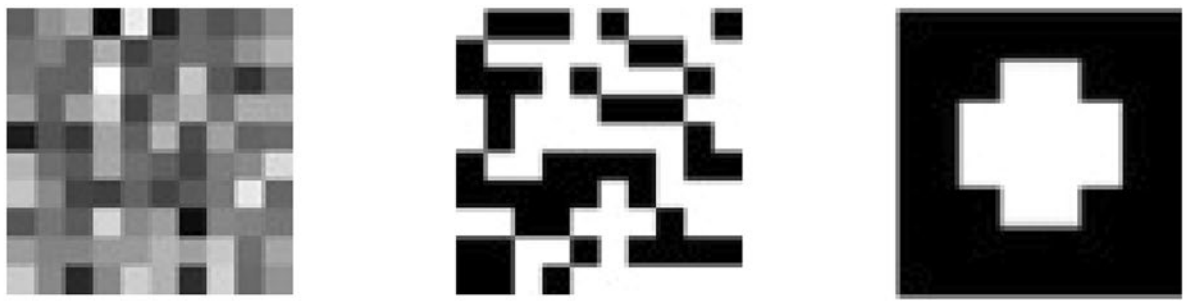

(a) Gaussian noise (b) Bernoulli noise (c) Binary mask

**Fig. 2** Comparison of Gaussian noise (left) with Bernoulli noise (center) and a binary mask (right), demonstrating how Bernoulli noise's discrete nature aligns more closely with the binary nature of segmentation masks, highlighting its potential for enhancing model performance in binary data segmentation

Here, $x_t$ is sampled from a Bernoulli distribution with probability parameter $(1-\beta_t)x_{t-1}+\beta_t/2$, where $\beta_t$ is indicative of a predetermined noise schedule that modulates the degree of noise incorporated at each individual timestep. The primary objective of the reverse process is to precisely restore the initial binary data from its noise-altered state via a reverse diffusion sequence, and it is defined by the distribution $p_\theta\left(x_{t-1}|x_t\right)$ as below:

$$p_\theta\left(x_{t-1}|x_t\right) = Bernoulli\left(x_{t-1};\Phi_\theta\left(x_t,t\right)\right). \tag{6}$$

In this context, $\Phi_\theta(x_t,t)$ represents the probability parameter of the Bernoulli distribution predicted by the neural network model. The model is trained to learn these parameters, allowing it to guide the procedure of denoising effectively.

Bernoulli diffusion is particularly advantageous for tasks involving discrete data, such as binary segmentation masks in medical imaging, as shown in Fig. 2. By tailoring the noise addition and removal processes to the nature of the data, Bernoulli diffusion enhances the model's ability to produce precise and reliable binary segmentation results, leveraging the benefits of diffusion processes in a discrete domain.

### 3.3 Diffusion for image segmentation

This diffusion process can be effectively used for image segmentation tasks, as introduced in the paper SegDiff [6], and subsequently extended to medical image segmentation in MedSegDiff [7]. It introduces a novel conditioning mechanism that intricately intertwines the denoising process with the input image. In this refined approach, the model operates by engaging with two primary inputs: the original input image $I_{in}$ and the noisy image $x_t$ at a given timestep $t$. These inputs are processed through image encoders and combined into a shared latent space. By conditioning the denoising process on both the input image's spatial information and the noisy image's state, the model represented as $\Sigma_\theta(x_t, I_{in}t)$, produces accurate segmentation masks by effectively utilizing the provided spatial context. The use of diffusion processes for image segmentation represents a significant advancement in leveraging generative models for practical applications in medical imaging and forms the basis of our work.

## 4 Methodology

This section delves into the methodological framework of Echo-DND, our novel diffusion-based approach for achieving superior left ventricular (LV) segmentation accuracy in echocardiograms (Fig. 3). Echo-DND stands at the confluence of generative modeling and precise medical imaging, harnessing the power of dual noise diffusion and multi-scale conditioning to navigate the intricacies of echocardiogram analysis.

### 4.1 Dual Noise diffusion strategy

Echo-DND departs from conventional diffusion models that solely rely on Gaussian noise for the forward diffusion process. Instead, it leverages a dual noise strategy, seamlessly integrating both Gaussian and Bernoulli noises. This strategic choice caters specifically to the binary nature of segmentation tasks like LV segmentation.

In Echo-DND, the initial noise is modeled as a combination of Gaussian and Bernoulli distributions:

$$\left(x_T\right) \sim (N\left(0, I_{n\times n}\right), Bernoulli(0.5)^{n\times n}) \tag{7}$$

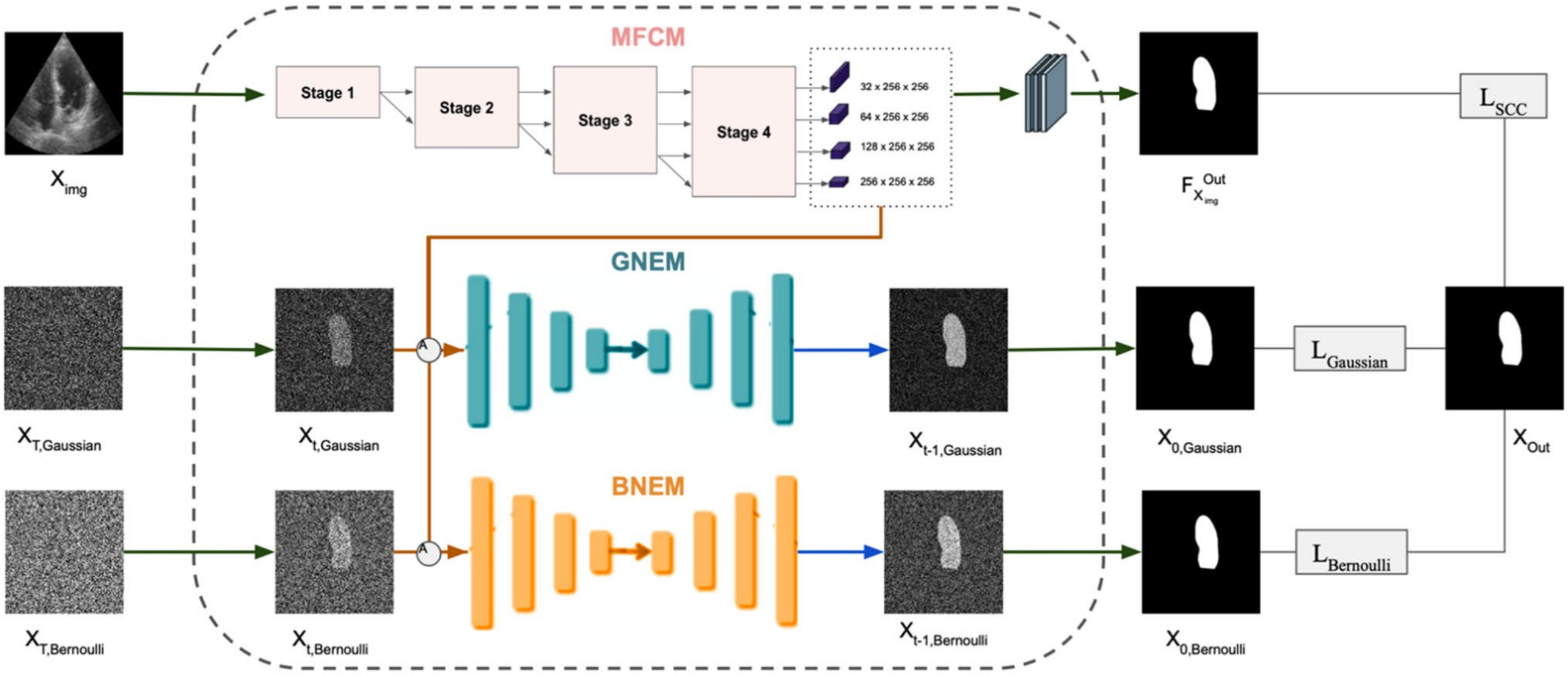

**Fig. 3** Comprehensive architecture of Echo-DND with Dual Noise Diffusion and Multi-Scale Fusion. This diagram encapsulates the Echo-DND model's intricate architecture, illustrating the integration of Gaussian and Bernoulli noise diffusion processes within a unified framework. The architecture highlights the twin noise estimation modules—GNEM and BNEM—alongside the Multi-scale Fusion Conditioning Module (MFCM), which collectively facilitates precise segmentation of echocardiograms. The diagram also delineates the loss calculation flow, showcasing the model's training methodology that leverages the spatial coherence calibration technique (LSCC) to refine segmentation accuracy

where $I_{n\times n}$ represents the identity matrix with dimensions corresponding to the echocardiogram image $X_{img}$ of size $n \times n$. This composite noise model allows the latent variables $x_{T,Gaussian}$ and $x_{T,Bernoulli}$ to represent the Gaussian and Bernoulli noise components, respectively. The primary objective of Echo-DND is to transform these noisy latent variables back to their clean states, $x_{0,Gaussian}$, and $x_{0,Bernoulli}$, thereby recovering accurate LV segmentation masks. Both of these masks are then fused together using the STAPLE algorithm [46] is to obtain the final segmentation mask.

Echo-DND employs a twin architecture for noise estimation, consisting of two modules: the Gaussian Noise Estimation Module (GNEM) and the Bernoulli Noise Estimation Module (BNEM). Both modules use a refined U-Net architecture [5], with an enhancement of a multi-head self-attention mechanism [31] at resolution $16 \times 16$ and $8 \times 8$, respectively. This enhancement is instrumental in capturing extensive long-range patterns and contextual intrinsic details within the image. Additionally, sinusoidal time embeddings are used to condition each block on the current timestep $t$, ensuring that temporal information is effectively integrated into the denoising process.

### 4.2 Multi-scale fusion-based conditioning

To enhance the accuracy and detail of echocardiogram segmentation, Echo-DND incorporates a Multi-scale Fusion Condition Module (MFCM), significantly improving upon conventional convolutional neural network (CNN) methodologies. The module draws inspiration from the High-Resolution Network (HRNet) [47] utilized in UTRNet [13]. It consistently upholds detailed high-resolution depictions across the entire network, crucial for capturing the intricate details of cardiac anatomy.

Echo-DND ensures that high-resolution feature maps are preserved at every stage of the network. Preserving high resolution is particularly critical in echocardiography due to often subtle texture differences and ambiguous endocardial borders that can be lost with excessive downsampling. Traditional CNNs often down-sample images to extract features, which can lead to a loss of spatial resolution and fine details. In contrast, the Multi-scale Fusion Conditioning module avoids this pitfall by sustaining multiple scales of resolution simultaneously. At each stage of the network, multiresolution feature streams are generated and maintained. These streams coexist and interact, allowing the network to enrich the feature maps with detailed, high-resolution information. This approach allows the model to capture both global structures and fine-grained details, essential for accurate segmentation of complex anatomical structures in echocardiograms. Specifically, the module generates semantically rich conditional features denoted as $F_X^i$ for the ith scale. During the noise estimation process, each layer of conditional features $F^i_{Ximg}$ is meticulously fused with the corresponding encoding features via a cross-attention mechanism. This enables the model to dynamically weigh the importance of different features according to their relevance to the current task. This approach significantly improves the model's proficiency in demarcating the contours of the LV with elevated accuracy, leading to more accurate and reliable segmentation results.

### 4.3 Training methodology

Our training methodology aligns with the principles of diffusion probabilistic models (DPMs) [5] and involves optimizing two core loss functions: $L_{\text{Gaussian}}$ and $L_{\text{Bernoulli}}$, corresponding to the Gaussian and Bernoulli noise diffusion components, respectively. The loss functions are defined as expectations over the ground truth data $x_0$ and uniformly sampled timesteps $t \in \{1, \dots, T\}$ (denoted by) $\mathbb{E}_{x_0,t}$:

$$L_{Gaussian} = \mathbb{E}_{x_0,t}\left[ \|\epsilon - \epsilon_\theta\left(\sqrt{\hat{\alpha}_t}x_0 + \sqrt{1-\hat{\alpha}_t}\epsilon, X_{img}, t\right)\|^2 \right] \tag{8}$$

$$L_{Bernoulli} = -\mathbb{E}_{x_0,t}\left[x_0 \log\left(\phi_\theta(x_t^B, X_{img}, t)\right)\right] + (1-x_0)\log\left(1-\phi_\theta(x_t^B, X_{img}, t)\right) \tag{9}$$

where $x_0$ represents the ground truth binary segmentation mask.$X_{img}$ denotes the input echocardiogram image, serving as conditioning information.$t$ is a timestep uniformly sampled from$\{1, \dots, T\}$. $x_t^B$ signifies the noisy segmentation mask at timestep $t$ under Bernoulli forward process.$\epsilon$ is the true Gaussian noise sampled from $N(0, I)$ used in the Gaussian forward process. $\hat{\alpha}_t$ is the cumulative product related to noise schedule $\beta_t$ as per standard DDPM notations [5]. $\epsilon_\theta$ is the parameterized function representing the GNEM , $\phi_\theta$ is the parameterized function representing the BNEM.

These loss functions measure the discrepancy between the predicted and true noise residuals at each timestep $t$ and minimizing them refines the prediction of the corresponding noise parameters.

Building upon the variational lower bound principle used in previous diffusion models [48], we also incorporate a Kullback–Leibler (KL) divergence term into our training paradigm. This term maintains the distributional integrity of the process, ensuring that our model's predictions adhere to the statistical properties of target noise distributions. Specifically, the KL divergence term encourages the learned distributions to remain close to their standard priors (standard Normal and uniform Bernoulli, respectively). This regularization acts as a constraint, potentially aiding training stability and improving the model's generalization capabilities by preventing overfitting to specific noise patterns encountered during training. For the Gaussian component, the KL divergence term $L_{\text{KL,Gaussian}}$ is formulated as:

$$L_{\text{KL,Gaussian}} = E_t\left[KL(N(\mu_\theta\left(x_t, t\right), \Sigma_\theta(x_t, t))||N(0, I))\right]. \tag{10}$$

For the Bernoulli component, the KL divergence term $L_{\text{KL,Bernoulli}}$ ensures adherence to a balanced Bernoulli distribution:

$$L_{\text{KL,Bernoulli}} = E_t\left[KL(Bernoulli(x_t^B, X_{img}, t))||Bernoulli(0.5))\right] \tag{11}$$

These terms ensure that Echo-DND's output distributions for both Gaussian and Bernoulli noises align with their respective standard priors.

Additionally, we incorporate pixel-wise spatial coherence calibration (SCC) into the training process. This involves a typical feedforward segmentation-style cross-entropy loss. The features extracted by the multi-scale fusion-based conditioning module are denoted as $F^o_{X_{img}}$ at the 0th scale. These features are passed through an FCN block and transformed to match the dimensions of the output segmentation mask, and this is denoted as $F^{out}_{X_{img}}$. The SCC loss, calculated as the Mean Squared Error (MSE) between the segmentation mask and the conditional features is given by:

$$L_{SCC} = E_{x_0}\|x_0 - F^{out}_{X_{img}}\|^2. \tag{12}$$

This SCC loss utilizes MSE as a direct and standard metric to enforce pixel-wise similarity between the conditioning module's high-resolution feature representation (after transformation) and the target ground truth segmentation map. By doing so, it explicitly guides the conditioning module to extract features pertinent to segmentation task and promotes spatial coherence in the features that condition the diffusion process, complementing the fine-grained refinement learned via the primary diffusion losses. The total loss function is a weighted combination of these terms:

$$L_{total} = \lambda_1 L_{Gaussian} + \lambda_2 L_{Bernoulli} + \lambda_3 L_{KL,Gaussian} + \lambda_4 L_{KL,Bernoulli} + \lambda_5 L_{SCC} \tag{13}$$

**Table 1** Comparison of CAMUS and EchoNet-Dynamic datasets

| Feature | CAMUS dataset | EchoNet-dynamic dataset |
|---|---|---|
| Source | Saint-Etienne' University Hospital | Stanford Health Care |
| Content-type | Echocardiographic images | Echocardiogram videos |
| Views | A2C and A4C | A4C |
| Number of samples | 500 patients (450 annotated) | 10,030 videos |
| Image dimensions | Variable | 112 × 112 pixels per frame |
| Data split | 400 training, 50 validation | 7465 training, 1277 validation |
| Format | Raw/mhd file format | RGB encoded, extracted from DICOM files |

In this formulation, $\lambda_1$, $\lambda_2$, $\lambda_3$, $\lambda_4$ and $\lambda_5$ are hyperparameters that control the relative importance of each loss term. In our experiments, the weight values were set as $\lambda_1 = \lambda_2 = 1$, $\lambda_3 = \lambda_4 = 0.01$, and $\lambda_5 = 0.1$.

## 5 Experiments and Results

### 5.1 Dataset

Our empirical investigations leveraged two eminent publicly available datasets: CAMUS and EchoNet-Dynamic (Ref. Table 1).

The CAMUS dataset [1], known as The Cardiac Acquisitions for Multi-structure Ultrasound Segmentation (CAMUS) dataset, is a comprehensive collection of clinical, echocardiographic images from 500 patients, curated at the University Hospital of St Etienne, France. This dataset is particularly notable for its reflection of real-world clinical scenarios, encompassing a wide array of cases, including those with ambiguous LV boundaries that pose significant segmentation challenges. Out of the total patient scans, 450 are annotated, providing a rich source of ground truth for model training and validation. For our study, we partitioned these annotated cases into a training set of 400 patients and a validation set of 50, ensuring a robust evaluation framework for Echo-DND's performance.

The EchoNet-Dynamic dataset [49], on the other hand, presents a voluminous repository of 10,030 apical four-chamber (A4C) echocardiogram videos, each meticulously labeled with LV tracings, ejection fraction, and volumetric measurements at the end-systole (ES) and end-diastole (ED). These videos, sourced from Stanford Health Care between 2016 and 2018, offer a granular view of cardiac function through 112 × 112 pixel frames extracted from DICOM files. Following the original dataset's partitioning scheme, we utilized 7465 videos for training and 1277 for validation, resulting in a total of 14,930 training images and 2554 validation images. This dataset's extensive size and detailed annotations make it an exemplary benchmark for assessing the capabilities of our Echo-DND model in segmenting dynamic cardiac structures.

### 5.2 Implementation details

The Echo-DND model was meticulously designed and implemented using PyTorch to ensure robust performance in echocardiogram segmentation. The input images were standardized to 256 × 256 pixels, maintaining the original aspect ratio through padding. To enhance model robustness against variations in clinical data and improve generalization, standard online data augmentation techniques were applied during training, including random horizontal flips, minor rotations, slight random scalings and random adjustments to brightness and contrast. We also introduced low-intensity Gaussian noise augmentation to reflect the inherent noise in ultrasound imaging. These augmentations were applied on-the-fly to each training sample, preserving data diversity while minimizing the risk of overfitting. Echo-DND's architecture utilizes 128 channels (C = 128) at its core to capture the complex features inherent in echocardiographic images. Due to the three extensive modules in it, the total number of parameters in the model is 238M. The training process was executed on an NVIDIA A100 40GB GPU, with a batch size of 8 over 200,000 steps. The AdamW optimizer was employed

**Table 2** CAMUS dataset-A comparative analysis of Echo-DND and SOTA models in Left Ventricular segmentation. Bold denotes the best value in each column

| Method | End-systolic | End-diastolic | Overall |
|---|---|---|---|
| U-Net (E) [11] | 0.927 | 0.951 | 0.939 |
| nnU-Net [19] | 0.945 | 0.939 | 0.942 |
| Unet++ (E) [20] | 0.951 | 0.955 | 0.953 |
| MFP-Unet [27] | 0.941 | 0.949 | 0.945 |
| ResUNet (E) [21] | 0.946 | 0.948 | 0.947 |
| ResDUnet [24] | 0.948 | 0.954 | 0.951 |
| DeepLabV3 [55] | 0.865 | 0.869 | 0.867 |
| TransUNet (E) [34] | 0.951 | 0.955 | 0.953 |
| SegFormer [35] | 0.952 | 0.957 | 0.955 |
| Swin Transformer [11] | 0.955 | 0.957 | 0.956 |
| SegDiff (E) [6] | 0.893 | 0.891 | 0.892 |
| EnsemDiff (E) [43] | 0.898 | 0.924 | 0.911 |
| BerDiff (E) [44] | 0.931 | 0.928 | 0.929 |
| MedSegDiff (E) [7] | 0.952 | 0.949 | 0.951 |
| **Echo-DND (Ours)** | **0.971** | **0.954** | **0.962** |

Boldface is used to highlight the best Dice score with a learning rate of $5 \times 10^{-5}$ to facilitate stable and efficient convergence. We currently employ DDIM [53] for sampling with 1000 diffusion steps, which translates to an inference time of around 2 min per image on an NVIDIA A100. Dice score was employed to assess segmentation accuracy, ensuring robust performance evaluation. This implementation framework establishes Echo-DND as a new benchmark in echocardiogram segmentation.

### 5.3 Results and discussion

In this section, we present a comprehensive evaluation of Echo-DND's performance through various experiments, benchmarking it against several state-of-the-art models in medical image segmentation and conducting detailed ablation studies to optimize its components. The performance of Echo-DND was rigorously compared with existing models on the CAMUS and EchoNet-Dynamic datasets. All models were trained on the respective training sets and validated on the validation sets, ensuring fair and unbiased comparisons. Echo-DND's results are derived from our own experiments, while the metrics for other models are sourced from existing literature or obtained using the standard settings of their open-source implementations (as indicated by "(E)" in Tables 2, 3).

The comparative analysis presented in Table 2 demonstrates Echo-DND's exceptional performance on the CAMUS dataset, achieving an overall Dice score of 0.962. This performance surpasses traditional architectures such as U-Net, ResUNet, and DeepLabV3. We attribute this gain, in part, to Echo-DND's architectural innovations: the Multi-scale Fusion Conditioning Module (MFCM) effectively preserves high-resolution spatial details often lost in standard CNN encoder-decoders, while the iterative refinement intrinsic to the diffusion process allows for more robust handling of inherent image noise and complex textures compared to single-pass segmentation. The proposed Echo-DND outperforms recent transformer-based models like TransUNet, SegFormer, and Swin Transformer. While transformers leverage powerful attention mechanisms for capturing global context, diffusion models like Echo-DND excel at generative refinement, learning to reverse a noise process to produce highly detailed segmentations. This iterative denoising, combined with Echo-DND's dual-noise strategy tailored for echo's characteristics appears particularly effective for modeling pixel-level uncertainties and achieving precise boundary delineation in this challenging modality. Furthermore, Echo-DND markedly exceeds the performance of other diffusion-based models, including SegDiff, BerDiff, and MedSegDiff. This highlights the efficacy of the proposed dual noise strategy; we hypothesize that synergistically combining Gaussian noise (adept at modeling continuous sensor/background variations) and Bernoulli noise (better aligned with the discrete nature of segmentation) affords greater adaptability and robustness to echocardiographic data compared to single-noise diffusion frameworks. The integration of MFCM conditioning and the spatial coherence calibration (SCC) loss further contributes by ensuring both rich contextual input and spatial consistency in the output masks.

On the EchoNet-Dynamic dataset, Echo-DND continued to exhibit cutting-edge performance, achieving an overall Dice score of 0.939 (Table 3). It is pertinent to note that performance scores across all models are generally lower on

**Table 3** ECHONET-Dynamic dataset- Demonstrating Echo-DND's cutting-edge performance against SOTA models. Bold denotes the best value in each column

| Method | End-systolic | End-diastolic | Overall |
|---|---|---|---|
| U-Net (E) [11] | 0.927 | 0.919 | 0.923 |
| nnU-Net (E) [19] | 0.931 | 0.926 | 0.929 |
| Unet++ (E) [20] | 0.935 | 0.932 | 0.934 |
| ResUNet [21] | 0.912 | 0.935 | 0.924 |
| DeepLabV3 [55] | 0.903 | 0.927 | 0.915 |
| ASPP-ED [51] | 0.923 | 0.940 | 0.933 |
| SimLVSeg-3D [52] | 0.916 | 0.937 | 0.929 |
| TransUNet [34] | 0.921 | 0.929 | 0.925 |
| SegFormer [35] | 0.925 | 0.931 | 0.928 |
| Swin Transformer [11] | 0.922 | 0.938 | 0.930 |
| TransBridge [33] | 0.902 | 0.930 | 0.916 |
| SegDiff (E) [6] | 0.871 | 0.891 | 0.881 |
| EnsemDiff (E) [43] | 0.879 | 0.903 | 0.891 |
| BerDiff (E) [44] | 0.911 | 0.919 | 0.915 |
| MedSegDiff (E) [7] | 0.914 | 0.929 | 0.921 |
| **Echo-DND (Ours)** | **0.946** | **0.932** | **0.939** |

Boldface is used to highlight the best Dice score

EchoNet-Dynamic compared to CAMUS, likely reflecting the increased challenge posed by EchoNet's inherently lower spatial resolution (112 × 112 vs. variable, often higher, in CAMUS) and potentially different noise characteristics. Despite these challenges, Echo-DND again surpasses established methods, including U-Net, ResUNet, and recent specialized approaches like ASPP-ED. It also outperforms contemporary transformer models like SegFormer and Swin Transformer, as well as other diffusion models such as MedSegDiff These results underscore Echo-DND's adaptability and precision across datasets with varying characteristics, reinforcing the robustness conferred by its core components (dual noise, MFCM, diffusion refinement) even on lower-resolution dynamic sequences.

Analyzing the performance nuances further, particularly on the CAMUS dataset, we observe an interesting pattern. While Echo-DND's end-diastolic segmentation score (0.954) is competitive yet slightly below the peak achieved by SegFormer (0.957), its end-systolic score (0.971) is substantially higher than all other reported methods. This exceptional performance during end-systole, a phase where the LV volume is minimal and segmentation becomes particularly challenging due to the smaller target size and potential shape complexities, strongly highlights Echo-DND's capabilities. We posit that this advantage stems from the interplay between the diffusion model's iterative refinement process, which allows meticulous delineation even for small structures, and the MFCM's capacity to preserve fine-grained spatial details throughout the network. This suggests Echo-DND offers a distinct advantage in scenarios demanding high precision on smaller or more intricate anatomical features, even if global context modeling by top transformers might yield marginal gains in less demanding phases like end-diastole. This capability underscores Echo-DND's potential as a robust clinical tool, particularly valuable where traditional methods falter.

Ablation studies revealed several critical insights into Echo-DND's individual components, as summarized in Table 4. These studies evaluated the impact of varying training steps, conditioning modules, loss functions, and noise types on segmentation accuracy. The optimal number of training steps was found to be 200,000 (Dice 0.962). This balance indicates that extensive training allows the model to refine its predictions accurately, but overtraining can lead to diminishing returns. Crucially, the custom MFCM conditioning module yielded significantly better results (0.962) compared to a standard U-Net conditioner (0.958), confirming its effectiveness in leveraging multi-scale spatial and contextual information. The comprehensive loss function, including the spatial coherence calibration term (LSCC), was crucial in achieving high segmentation accuracy. This term helps maintain the spatial integrity of the segmentation masks, reducing noise and ensuring consistency in the predictions. Furthermore, the combination of Gaussian and Bernoulli noise terms provided the best performance (0.962) compared to using either noise type alone (0.949 for Gaussian, 0.954 for Bernoulli). This result highlights how the synergistic interplay between continuous and discrete noise distributions helps the model learn the diverse noise patterns during training. This diversity likely helps the model generalize better to various types of noise and artifacts present in echocardiographic images.

**Table 4** Ablation insights-Fine-Tuning Echo-DND's architecture for optimal segmentation accuracy

| Ablation study component | Variations | Dice score |
|---|---|---|
| Number of training steps | 50k | 0.927 |
| | 100k | 0.958 |
| | 150k | 0.961 |
| | 200k | 0.962 |
| | 250k | 0.960 |
| | 300k | 0.959 |
| Conditioning module | U-Net | 0.958 |
| | MFCM | 0.962 |
| Loss function | $L_G + L_B$ term | 0.951 |
| | $+KL$ terms | 0.955 |
| | $+L_{SCC}$ term | 0.962 |
| Noise term | Only Gaussian | 0.949 |
| | Only Bernoulli | 0.954 |
| | Both | 0.962 |

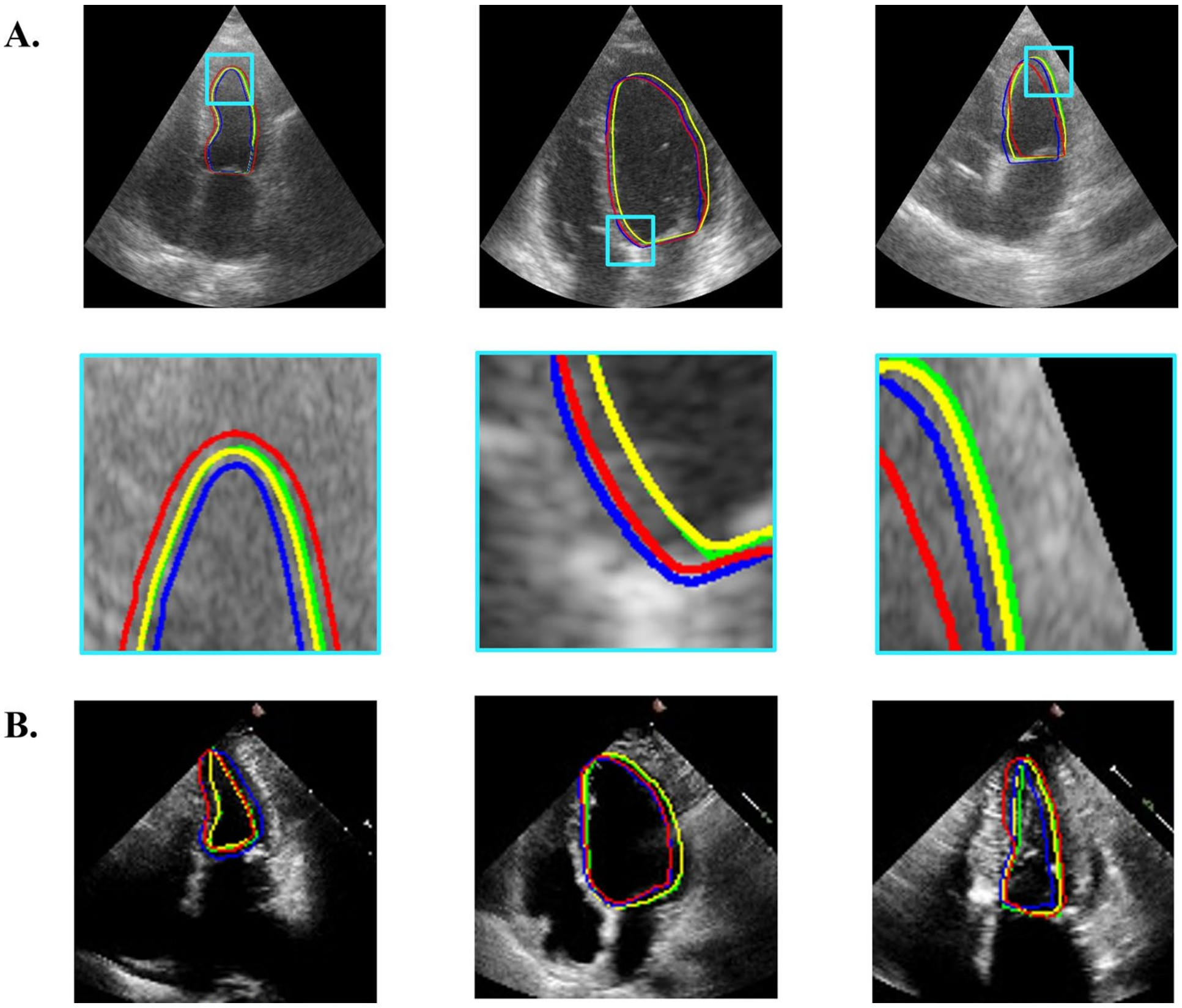

**Fig. 4** Comparative visualization of left ventricle segmentation on representative samples from CAMUS and EchoNet-Dynamic datasets. The contours represent: ground truth (green line), Echo-DND (yellow line), U-Net (red line), and MedSegDiff (blue line). These images exhibit common clinical challenges such as low contrast and boundary ambiguity. **A)** To showcase precision on difficult boundary segments, zoomed-in insets are provided for the CAMUS examples. **B)** Due to the inherent low resolution (112 × 112 pixels) of the EchoNet-Dynamic dataset, which limits the clarity gained from magnification, corresponding insets are not shown. This qualitative analysis illustrates Echo-DND's generally superior alignment with the ground truth compared to the baseline methods across diverse data characteristics

A qualitative analysis, presented in Fig. 4, offers further insight into model performance using challenging clinical examples from the CAMUS and EchoNet-Dynamic validation sets. These cases were chosen as representative examples of common echocardiographic difficulties, such as suboptimal image quality, low myocardium-blood pool contrast, and indistinct endocardial borders, particularly those in apical or lateral regions. The visualizations demonstrate that Echo-DND (yellow contours in Fig. 4) generally achieves superior alignment with the ground truth compared to the U-Net and MedSegDiff baselines, even amidst these challenges. For the higher-resolution CAMUS images, the included zoomed-in views specifically highlight Echo-DND's enhanced precision in delineating difficult boundary segments. This visual evidence corroborates the quantitative results, underscoring Echo-DND's robustness and accuracy on clinically relevant, non-trivial cases.

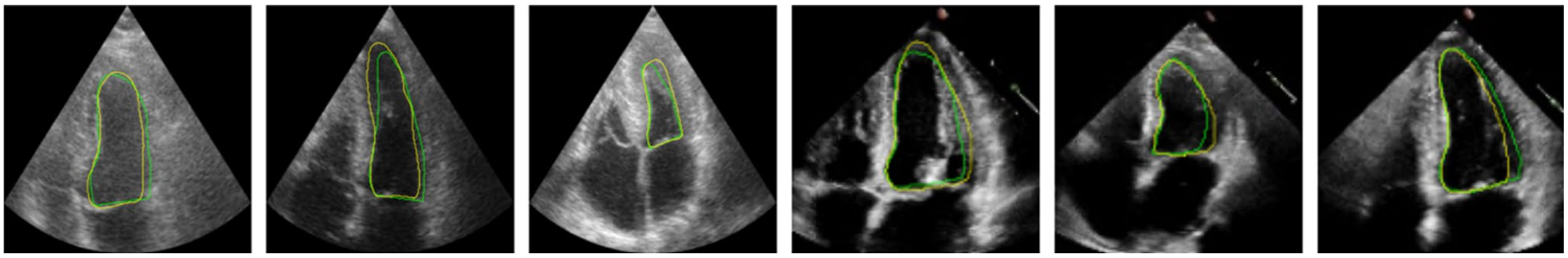

**Fig. 5** Illustration of Echo-DND segmentation limitations on particularly challenging edge cases. These images reveal instances where Echo-DND's segmentation (line yellow contour) deviates significantly from the ground truth (line green contour). Such failures typically occur under conditions of severe imaging artifacts, extremely low contrast, substantial signal dropout (affecting regions like the apex or lateral wall), significant acoustic shadowing, or complex anatomical interference (e.g., from trabeculations or papillary muscles). These scenarios represent the current limitations faced by automated segmentation methods when dealing with the most difficult clinical image data

While Echo-DND demonstrates strong performance overall, Fig. 5 focuses on its limitations by showcasing segmentation results on particularly demanding edge cases. In these scenarios—which may feature severe signal dropout (e.g., apical or lateral wall invisibility), extremely low contrast rendering boundaries nearly imperceptible, significant acoustic shadowing, or complex interference from intracardiac structures like dense trabeculations or papillary muscles—Echo-DND's segmentation can deviate noticeably from the ground truth. Analyzing these failure modes, which push the boundaries of current automated methods, is crucial for understanding present limitations and directing future research towards developing algorithms with even greater resilience to the full spectrum of clinical image variability and quality.

### 5.4 Future work

The promising results of Echo-DND indicate potential for broader applications in medical imaging. Future investigations will aim to extend Echo-DND's capabilities beyond echocardiogram segmentation to a wider range of medical imaging tasks. We will experiment with incorporating multiple noise types, such as gamma and Poisson, tailored to specific imaging modalities. Furthermore, extending the framework to leverage temporal information inherent in echocardiography sequences is a key direction; incorporating intermediate frames or recurrent structures within the diffusion process could enhance segmentation consistency and temporal smoothness across the cardiac cycle. Efforts will also focus on improving the speed of the diffusion-based process, addressing a significant barrier to clinical adoption. Exploring advanced sampling algorithms (e.g., DPM-solver, DPM++) and optimized step scheduling remains pertinent. Additionally, investigating model compression techniques—such as knowledge distillation, potentially utilizing efficient teacher-student architectures [54], alongside methods like network pruning and quantization—represents a crucial avenue for enhancing computational feasibility and deployment efficiency. Collectively, these advancements aim to further refine Echo-DND's performance, versatility, and clinical applicability across a spectrum of medical image analysis tasks.

## 6 Conclusion

This research introduces Echo-DND, an innovative dual-noise diffusion model tailored for precise left ventricle segmentation in echocardiograms. Through comprehensive evaluations, Echo-DND has demonstrated superior performance over both traditional and contemporary state-of-the-art models on the CAMUS and EchoNet-Dynamic datasets, achieving high Dice scores (0.962 and 0.939, respectively) and excelling particularly in challenging end-systolic segmentation. The model's success is underpinned by key architectural innovations: the synergistic dual-noise (Gaussian and Bernoulli) strategy effectively models complex noise characteristics and the binary nature of the task; the custom multi-scale fusion conditioning module preserves critical spatial details; and the comprehensive loss function ensures both distributional integrity and spatial coherence. Rigorous ablation studies validate the contribution of each component. Echo-DND's robust performance on diverse datasets highlights its potential not only as a new benchmark for echocardiogram segmentation but also for broader applications in medical imaging. Its potential integration into clinical workflows could facilitate faster, more reproducible quantitative cardiac analysis, serving as a valuable decision support tool for clinicians. Future research will focus on extending the Echo-DND framework, potentially adapting it to other imaging modalities where diffusion models are gaining traction, such as cardiac MRI or CT, while concurrently addressing computational efficiency for enhanced clinical translation. In summary, Echo-DND represents a significant advancement in medical image segmentation, offering a powerful solution for accurate and reliable echocardiogram analysis. Its success paves the way for future innovations in medical imaging, promising improved diagnostic capabilities and enhanced patient care.

**Acknowledgements** We extend our gratitude to the IIT Delhi High Performance Computing (HPC) facility for providing the GPU resources essential for our experiments. We are grateful to the Yardi School of Artificial Intelligence for their valuable assistance throughout this project. Special thanks to our mentors and collaborators for their invaluable guidance and support.

**Author contributions** AR: Data Curation, Software, Conceptualization, Methodology, Investigation, Writing—Original Draft; BKV: Data Curation, Conceptualization, Methodology, Investigation, Writing—Review & Editing. MKR: Conceptualization, Methodology, Investigation, Writing—Review & Editing. ASR: Conceptualization, Methodology, Investigation, Writing—Review & Editing.

**Funding** This research was supported by the Tower project (RP04636N, PI: ASR) at IIT Delhi. We extend our gratitude to the IIT Delhi High Performance Computing (HPC) facility for providing the GPU resources essential for our experiments. We are grateful to the Yardi School of Artificial Intelligence for their valuable assistance throughout this project. Special thanks to our mentors and collaborators for their invaluable guidance and support.

**Data availability** This study has been conducted using publicly available data and it can be accessed using the provided URLs: 1. CAMUS dataset: https://www.creatis.insa-lyon.fr/Challenge/camus/ and 2. https://stanfordaimi.azurewebsites.net/datasets/834e1cd1-92f7-4268-9daa-d359198b310a. We preprocessed the data.

**Code availability** To further the advancement of research in this field, we will make our code and model publicly available on GitHub.

## Declarations

**Ethics approval and consent to participate** Not applicable.

**Consent to publication** Not applicable.

**Competing interests** The authors declare no competing interests.